\documentclass{article}

\usepackage[final,nonatbib]{tccml_neurips_2020}

\usepackage[utf8]{inputenc} 
\usepackage[T1]{fontenc}    
\usepackage{hyperref}       
\usepackage{url}            
\usepackage{booktabs}       
\usepackage{amsfonts}       
\usepackage{nicefrac}       
\usepackage{microtype}      

\usepackage{subfigure}
\usepackage{graphicx}
\usepackage{cite}
\usepackage{graphicx, multicol, latexsym, amsmath, amssymb}
\usepackage{float}
\usepackage{siunitx}

\title{A Temporally Consistent Image-based Sun Tracking Algorithm for Solar Energy Forecasting Applications}

\author{%
  Quentin Paletta \\
  Department of Engineering\\
  University of Cambridge\\
  Cambridge, UK \\
  \texttt{qp208@cam.ac.uk} \\
    
    \And
    Joan Lasenby \\
    Department of Engineering\\
    University of Cambridge\\
    Cambridge, UK \\
    \texttt{jl221@cam.ac.uk} \\
  
}

\begin{document}

\maketitle

\begin{abstract}
 Improving irradiance forecasting is critical to further increase the share of solar in the energy mix. On a short time scale, fish-eye cameras on the ground are used to capture cloud displacements causing the local variability of the electricity production. As most of the solar radiation comes directly from the Sun, current forecasting approaches use its position in the image as a reference to interpret the cloud cover dynamics. However, existing Sun tracking methods rely on external data and a calibration of the camera, which requires access to the device. To address these limitations, this study introduces an image-based Sun tracking algorithm to localise the Sun in the image when it is visible and interpolate its daily trajectory from past observations. We validate the method on a set of sky images collected over a year at SIRTA's lab. Experimental results show that the proposed method provides robust smooth Sun trajectories with a mean absolute error below 1\% of the image size.
\end{abstract}

\vspace{-0.1\baselineskip}
\section{Introduction}
\vspace{-0.1\baselineskip}

The share of renewables in the global power mix is growing rapidly, with solar energy accounting for the largest growth~\cite{InternationalEnergyAgencyIEA2018}. However, the uncertainty and inherent variability of electricity production from photovoltaic panels is a major challenge for grid operators. Power system constraints and demand-supply balancing are among factors that currently limit its integration into the energy mix~\cite{Ela2013}. Improving solar electricity generation forecasting at different time-scales would mitigate these effects by facilitating battery, fossil fuel backup or grid management~\cite{Inman2013, West2014a, Antonanzas2016}. Beside statistical approaches, methods based on hemispherical sky cameras have fostered increased interest for short-term irradiance prediction. They provide high spatiotemporal resolution forecasts able to anticipate rapid irradiance fluctuations by capturing cloud cover changes over a solar facility~\cite{Kuhn2018a}. Because most of the incoming radiation originates from direct sunlight, a cloud hiding the Sun might cause a significant 80\% power drop within a minute. Consequently, short-term irradiance forecasting methods based on sky images largely use the position of the Sun in the image as a reference to predict the impact of the cloud cover dynamics on radiation changes~\cite{Chu2013a, Bernecker2014, Bone2018}.

Tracking the Sun position in the image is generally achieved by additional Sun tracking devices or through a well defined approach based on the relative position of the observer on Earth relative to the Sun~\cite{Reda2004, Blanc2012}. Following this procedure, the angular position of the Sun is accurately estimated with an uncertainty of $\pm0.0003^{\circ}$~\cite{Reda2004}. The main drawback of these methods is the translation of the angular coordinates into pixels, which must take into account the position and orientation of the camera and the strong distortion of the fish-eye lens. This can be a serious obstacle for researchers willing to use existing publicly available datasets~\cite{sirta, Pedro2019}, thus penalising open research in irradiance forecasting. Alternatively, image-based Sun tracking methods have been proposed, but they only work when the Sun is visible~\cite{Chu2016, Wei2016}, which limits their applicability to solar radiation applications.

To overcome this difficulty, we propose an alternative data-driven method based on the analysis of sky images only. From the observation that pixels corresponding to the Sun appear saturated in sky images, we robustly estimate the Sun's position in the image when it is visible. Following this we take advantage of the smooth trajectory of the Sun in the image from one minute to another, but also from day to day, to predict its daily trajectory from past observations only. We present quantitative and qualitative results of this approach and hope that future works will aim at validating the technique by comparing it to traditional methods.

\paragraph{Dataset}
The dataset used in this study was kindly shared by SIRTA's lab~\cite{sirta}. It is composed of several years of ground-taken sky images with a temporal resolution of 2 min. At each time step, two captures of the cloud cover are taken with a long (1/100 sec) and a short exposure (1/2000 sec). The only preprocessing applied to images is an azimuthal equidistant projection aiming at diminishing the lens distortion (see Appendix A).

\vspace{-0.1\baselineskip}
\section{Methods}
\vspace{-0.1\baselineskip}

The proposed framework is composed of three steps. Firstly, a binary classifier sorts images into two groups, `visible Sun' or `hidden Sun', based on the relative intensity of pixels in the image. If the Sun is visible, its position is estimated from its corresponding saturated pixels. Finally, the trajectory of the Sun over a day is interpolated from observations from previous days using supervised learning.

\paragraph{Binary classification}

\begin{figure}
  \centering
  \begin{minipage}[b]{0.32\textwidth}
    \includegraphics[width=\textwidth]{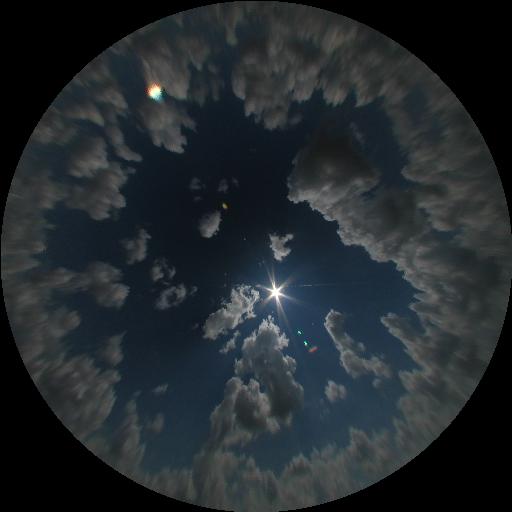}
    \label{fig:img_sh_2018_7_3_13_0}
  \end{minipage} 
  \begin{minipage}[b]{0.32\textwidth}
    \includegraphics[width=\textwidth]{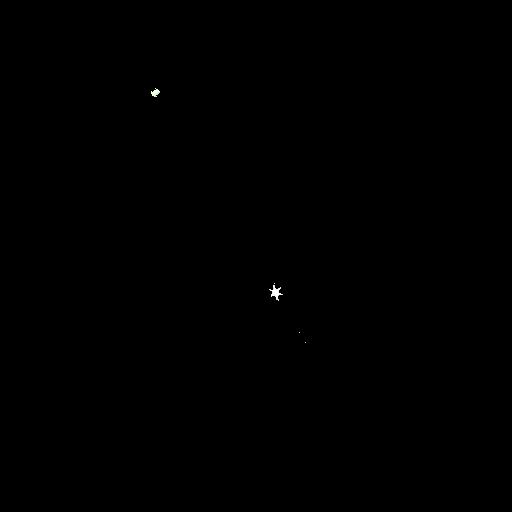}
    \label{fig:img_sun_204_2018_7_3_13_0}
  \end{minipage}
  \begin{minipage}[b]{0.32\textwidth}
    \includegraphics[width=\textwidth]{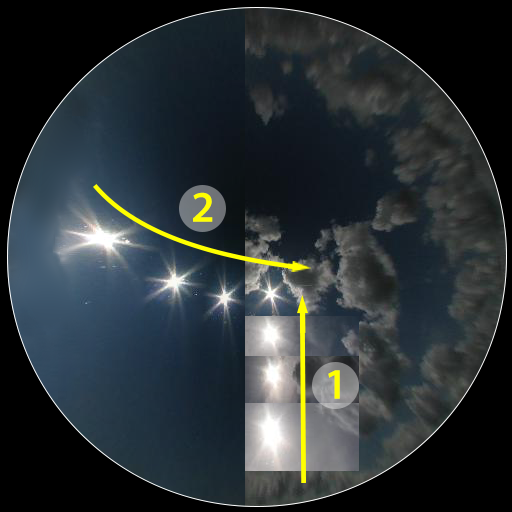}
    \label{fig:day_overlay}
  \end{minipage}
  \vspace{-1\baselineskip}
 \caption{From left to right: an image of the sky taken with a short exposure; the blue channel of the short exposure image once passed through a high intensity threshold (saturated areas are caused by direct sunlight and induced flare); a series of inlets showing the position of the sun extracted from consecutive sky images taken across (1) months (horizontally: March, April, May, June) and (2) hours of the same day (vertically: 6:00, 8:00, 10:00, 12:00). This superimposition highlights the smoothness of the Sun trajectory over time.}
 \label{fig:thresholded_image}
\end{figure}

The suggested classification decision is based on the maximal intensity of pixels in the images (Equation~\ref{equ:classification}): the Sun is assumed to be visible if the highest intensity pixel $\text{I}_{max}$ is above a percentage $p$ of the theoretical max $\text{I}_{theoretical \; max}$. Often available in sky image datasets\cite{sirta, Zhang2018}, images taken with a short exposure offer a higher contrast between the Sun and the background, while keeping the saturated area narrow, facilitating accurate Sun localisation (See Figure~\ref{fig:thresholded_image}). To simplify the processing of images, the threshold method was applied to the blue channel of the short exposure image, which offers the highest contrast.\footnote{The initial inclination was to use a Machine Learning-based classifier for this step. But given the impressive accuracy levels of the rule-based classifier that is implemented, using Machine Learning here was recognised as a computationally expensive choice with minimal accuracy gain expectations.}

\vspace{-0.3\baselineskip}

\begin{equation}
  \text{Sun is visible} \;\; \Leftrightarrow \;\; \text{I}_{max} > p \; \text{I}_{theoretical \; max} \;\; [\text{here} \;\; p = 0.99 \; \text{and} \; \text{I}_{theoretical \; max} = 255]
  \label{equ:classification}
\end{equation}

\paragraph{Sun localisation}

When the Sun is visible, the relative brightness of its corresponding pixels is used to guide its localisation in the image. As we can see in the second panel of Figure~\ref{fig:thresholded_image} however, induced flare can be of high intensity too. To discard this flare from the estimation of the Sun position, one can play with the size difference between the solar region and other saturated areas caused by artefacts. Given the higher number of pixels on the mode corresponding to the Sun, we first take the median position of the saturated pixels to robustly find an approximate position of the sun, before passing a Gaussian filter centred on it with a well tuned standard deviation to only retain the relevant saturated pixels matching the Sun while ignoring the flare. The median of the remaining pixel positions defines the estimated position of the Sun in the image. Similar approaches have been introduced by~\cite{Chu2016, Wei2016}.

\paragraph{Trajectory prediction} 
The next step of the tracking algorithm is to model the dependency of solar coordinates $x$ and $y$ in the image with time (days and minutes) to remove outliers and determine the position of the Sun when it is not visible, e.g. hidden by a cloud. The proposed method to estimate the trajectory of the Sun over a day using supervised learning, highlighted on the third panel of Figure~\ref{fig:thresholded_image}, is divided into two steps detailed below. Firstly, the Sun position is obtained from a training set of previous days' observations for every minute of the given day. An observation is classified as an outlier and removed from the set if the distance between the estimate from the Sun localisation algorithm and the prediction from previous observations is larger than 20 pixels (about 4\% of the size of the image). Secondly, a smooth trajectory of the Sun over that day is modelled from the resulting minute-by-minute estimates.

For each minute $m$ of a day $d$, a regularised polynomial regression or Ridge polynomial regression, is used to estimate the position of the Sun $\hat{x}_m^d$ (or $\hat{y}_m^d$) from previous days' observations of the visible Sun at the same minute $\text{x}_m^{d-1}$, $\text{x}_m^{d-2}$, ... $\text{x}_m^{d-N}$. In practice, a linear combination of the powers of the day $d_m^{i}$ ($0<i \leq I$) is used to model the dependency of $x_m^d$ with $d_m$ (see Equation~\ref{equ:ridge_1}).

\vspace{-1\baselineskip}

\begin{equation}
  \hat{x}_{d,m} = \sum_{i=0}^{I} \beta_i (d_m)^{i} = f(d_m) \quad \text{here} \quad I=4
   \label{equ:ridge_1}
\end{equation}

\vspace{-0.3\baselineskip}

To best fit the model to previous observations, the set of coefficients $\{ \beta_i \}$ is determined by minimising the quadratic difference between the observations $\{x_m^{d-n} \; with \; 0<n \leq N \}$ and their corresponding estimates $\{ \hat{x}_m^{d-n} \; with \; 0<n \leq N \}$. The trade-off between bias and variance is set through a regularisation parameter $\alpha_1$ (here $\alpha_1 = 0.01$) to avoid overfitting. An estimate is retained if it is obtained from at least four observations and if the last one occurred in the last ten days.

\vspace{-1\baselineskip}

\begin{equation}
  \hat{\beta} = \underset{\beta \in \mathbb{R}^\text{I+1}}{\operatorname{argmin}} \sum_{n=1}^{N} ||x_{d-n, m} - \hat{x}_{d-n, m}||^2_2 + \alpha_1 ||\beta||^2_2 \quad \text{here} \quad N = 60
  \label{equ:ridge_2}
\end{equation}

A smooth trajectory of the Sun, which is robust to outliers, is obtained through a second Ridge regression fitting the daily position estimates $\hat{x}_{d, m}$ of a given day from the previous regression. Equation~\ref{equ:ridge_3} presents the polynomial linear regression modelling the coordinate dependency with the 1440 minutes $m$ of a day. Following the same procedure, the vector $\hat{\lambda}$ is obtained through Equation~\ref{equ:ridge_3}. The regularisation parameter $\alpha_2$ is set to $10^{-7}$.

\vspace{-1\baselineskip}

\begin{equation}
  X_{d, m} = \sum_{i=0}^{I=4} \lambda_{d,i} \: m^{i} = g_d(m)  \; \; \text{with} \quad \hat{\lambda} = \underset{\lambda \in \mathbb{R}^\text{I+1}}{\operatorname{argmin}} \sum_{m=1}^{1440} ||\hat{x}_{d, m} - X_{d, m}||^2_2 + \alpha_2 ||\lambda||^2_2
  \label{equ:ridge_3}
\end{equation}

Depending on the application, diminishing the weight of the regularisation in the regressions will provide a closer fit to the data. However, this could penalise trajectory based approaches (optical flow, cloud tracking~\cite{Marquez2013}, the sector-ladder method~\cite{Quesada-Ruiz2014, Bone2018}, etc), which require spatial consistency between frames.

\vspace{-0.1\baselineskip}
\section{Experiments}
\vspace{-0.1\baselineskip}

\begin{figure}[ht]
\centering    
\includegraphics[width=0.95\textwidth]{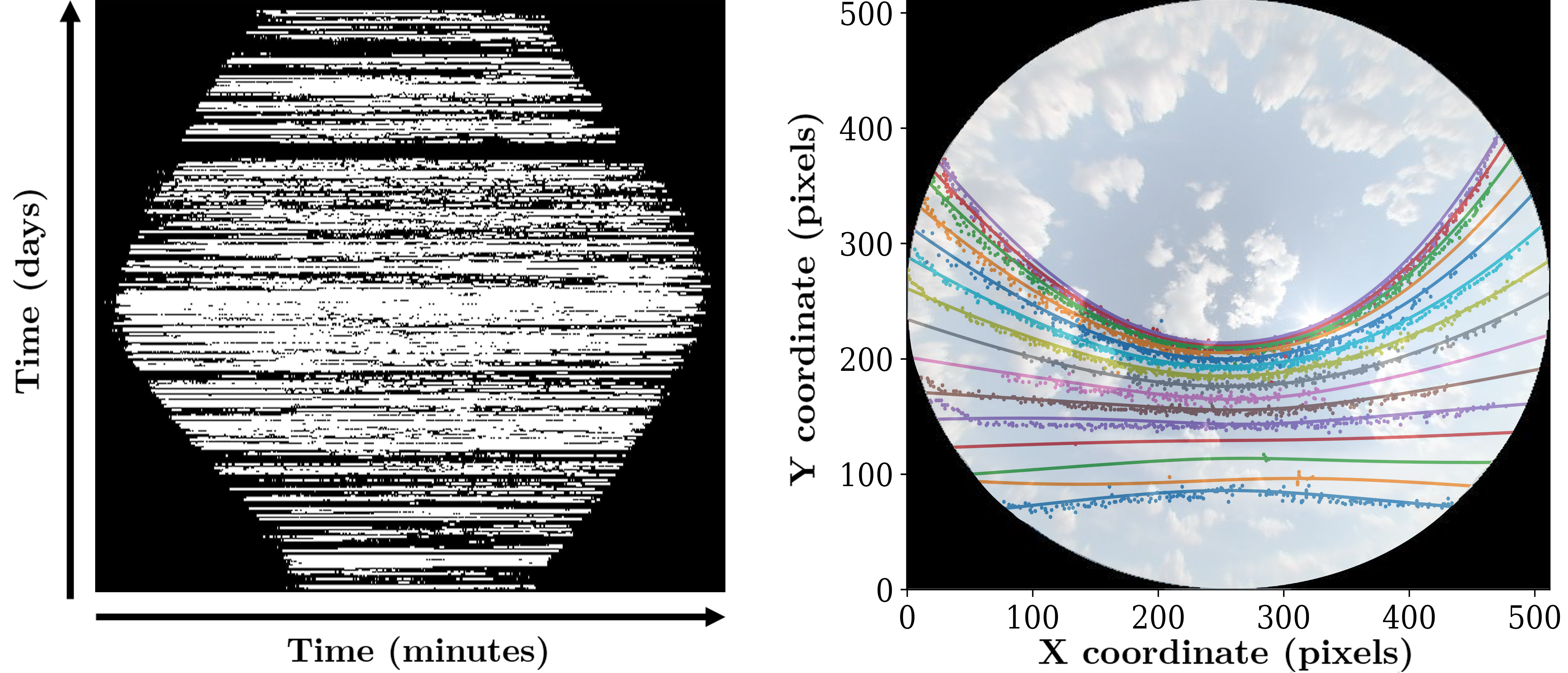}
\caption{From left to right: feature map showing the minutes (from 4:00 to 20:30 here) when the Sun was visible (white on the Figure) over 330 days (2017); position of the Sun in the sky for different days of the year (the 15 curves of the Figure correspond to 15 days sampled every 10 days from January to June 2017). Points correspond to observations of the Sun and curves to Sun trajectories over a day predicted by the algorithm from past observations.}
\label{fig:trajectory}
\end{figure}

The experiments presented in this work are conducted using 11 month of data collected in 2017~\cite{sirta}. The left panel in Figure~\ref{fig:trajectory} shows the distribution of visible Sun observations resulting from the classification algorithm. Large black stripes correspond to missing data (15 days in September).

The classification rule was evaluated on 360 manually labelled samples taken on a broken-sky day. Table~\ref{classification_score} shows that the algorithm reaches an accuracy of about 94\% with a precision of 98\%. Although straightforward, this decision rule appears reliable. Most errors occur in low Sun conditions, thus low irradiance conditions (False Negatives) or when the Sun is covered by a cloud, but strongly lights its edge, hence its position in the image is still approximately visible despite some inaccuracy (False Positives). One could further improve the method by training a deep classifier or a Support Vector Machine algorithm on a labelled dataset. 

\vspace{-0.6\baselineskip}

\begin{table}[H]
  \caption{Performance of the binary classification rule evaluated on 360 samples (Visible Sun: 240, Hidden Sun: 120) over a broken-sky day (02/06/2018)}
  \label{classification_score}
  \vspace{0.2\baselineskip}
  \centering
  \begin{tabular}{cccc}
    \toprule
    Accuracy & Precision & Recall & $\text{F}_1$ score \\
    \midrule
    94\% & 98\%  & 94\% & 96\% \\
    \bottomrule
  \end{tabular}
\end{table}

\vspace{-1\baselineskip}

The trajectory of the Sun over a year resulting from the Sun tracking algorithm is shown in Appendix B and in Figure~\ref{fig:trajectory}. Despite some minor discrepancies, the two-step modelling is robust to outliers and provides smooth trajectories even when the amount of data is limited like in the winter (Y below 150 pixels in Figure~\ref{fig:trajectory}). The largest inaccuracies occur when the Sun is close to the horizon, in the morning or in the evening. In these conditions however, the level of solar radiation is low or easily predictable.

\vspace{-0.2\baselineskip}

To contrast results, the size of the circumsolar area in undistorted images was measured along the trajectory of the Sun over a day in August. Its largest diameter ranged from 7 pixels in the middle of the day to 30 pixels in low Sun conditions (morning and evening). In comparison, the mean deviation between the final trajectory estimates and the visible Sun observations is about 4.7 pixels (0.91\% of the image size). Although not negligible, the error on the Sun trajectory is acceptable for most forecasting applications given that the size of clouds in the image are comparatively larger. In comparison, \cite{Chu2016} reported an average Sun location error based on its Sun tracking method of 3.71\% of the image, i.e. about 19 pixels here. Illustration of predictions by the Sun tracking algorithm are presented in Appendix C.

\vspace{-0.1\baselineskip}
\section{Discussion \& Conclusion}
\vspace{-0.1\baselineskip}

We propose in this paper a novel approach combining hand-crafted feature extraction and data-driven modelling to estimate the Sun trajectory in a sky image based on images only to overcome the difficulties and limitations of traditional methods. Results show that the model is able to provide smooth future trajectory estimates, which are robust to outliers and missing data. Future work will aim at better identifying incorrect estimates of the Sun position prior to data fitting. We hope that the method will be compared to existing approaches and applied to long exposure images to broaden its impact further.

\newpage

\section*{Broader Impact}


The percentage of solar energy has constantly been growing in the energy mix over recent years. As its penetration rises further, it is of paramount importance to obtain accurate forecasts of solar energy for integrating it effectively into the grid. By anticipating peaks and drops of a solar facility output, we can better mitigate the impact of its variability through battery, fossil fuel backup or grid management~\cite{Ela2013} and, consequently, contribute to an increase in both affordability of solar energy (which converts to the cost of energy for the end-users) and profitability of solar facilities. Therefore, improved solar energy forecasting would foster energy supply decarbonisation by facilitating effective integration of higher levels of solar resource into the grid, thereby reducing the total percentage of fossil fuels in the energy mix. Besides satellite imagery, ground observations of the cloud cover dynamics from fish-eye cameras is a promising approach to high spatiotemporal resolution forecasting. Capturing the changing structure of clouds can be used to predict the irradiance map over a solar facility and its corresponding future power output~\cite{Kuhn2018a}.

Despite the increasing availability of open access datasets~\cite{sirta, Pedro2019}, there are still inherent limitations to the application of common solar prediction methods to such datasets. For instance, given that the majority of solar radiation originates directly from the Sun, forecasting methods often analyse the displacement of clouds relative to the Sun's position in the image to predict the future incoming solar flux~\cite{Chu2013a, Bernecker2014, Quesada-Ruiz2014, Bone2018}. However, using this information requires access to the camera to calibrate its lens and evaluate the orientation / position of the device. For this reason, most publications to date apply methods of interest on datasets associated with the given study. In addition, to the best of our knowledge, the position of the Sun in sky images has not been used in previous studies applying Deep Learning to irradiance forecasting~\cite{Zhang2018, Sun2018, Sun2018a, Siddiqui2019, Zhao2019, Guen2020a}. Although there were attempts to base the Sun tracking on sky images, the method only works when the Sun is visible and does not provide consistent Sun trajectories~\cite{Chu2016, Wei2016}, which are for instance required for techniques based on cloud tracking, e.g.~\cite{Quesada-Ruiz2014, Chow2015}.

To address these obstacles, we propose a tracking method relying only on sky images to provide a temporally consistent Sun trajectory. The technique achieves a reliable performance from less than a week's worth of observations, and can thus be applied to small datasets or datasets collected by recently established solar facilities. By simplifying the implementation of irradiance forecasting methods, this study will stimulate open research in the field, at the same time offering effective ways for large scale deployment of solar forecasting systems without the need for expensive instrument installations.

Alongside the tangible benefits of the proposed method, inaccuracies in power prediction are a potential risk. These include an incorrect estimation of the Sun trajectory, affected by lack of robustness to outliers or a displacement of the camera making past observations unreliable. This, in turn, may cause financial penalties to producers, grid management issues, and possibly electric power shortages. Opportunities to increase the reliability of the method range from improved outlier detection to the implementation of alternative modelling techniques. In addition, its application to common long exposure images would automatically broaden its impact. We also hope that assessing the image-based Sun tracking with traditional approaches will bring additional insights to overcome its limitations, but also further demonstrate its reliability.

\begin{ack}

The authors acknowledge SIRTA for providing the sky images used in this study. We thank Sakshi Mishra and Guillaume Arbod for their guidance and valuable advice. This research was supported by ENGIE Lab CRIGEN, EPSRC and the University of Cambridge.

\end{ack}


\newpage

\bibliography{library}{}
\bibliographystyle{unsrt}

\small

\newpage

\section*{Appendix A: Preprocessing}
\label{preprocessing}

\paragraph{Undistortion}

\begin{figure}[H]
  \centering
  \begin{minipage}[b]{0.38\textwidth}
    \includegraphics[width=\textwidth]{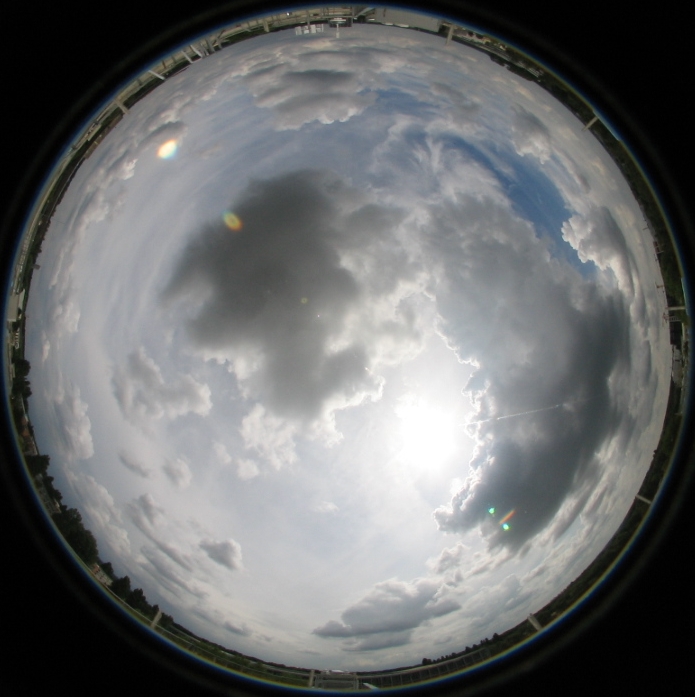}
    \label{fig:20180705132800_01_raw}
  \end{minipage} 
  \quad \quad \quad \quad
  \begin{minipage}[b]{0.38\textwidth}
    \includegraphics[width=\textwidth]{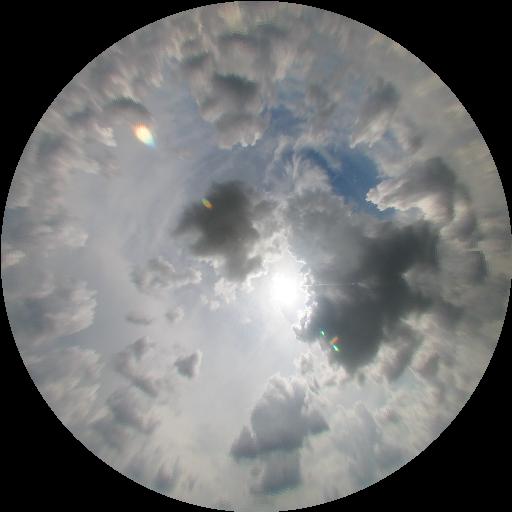}
    \label{fig:20180705132800_01_stereo}
  \end{minipage}
 \caption{Raw image (left) and its corresponding undistorted image (right) following an equidistant projection. The cloud cover is projected onto a horizontal plane to retrieve cloud shape and trajectory consistency between frames (05/07/2018, 13:28)}
 \label{fig:undistortion}
\end{figure}

Images taken by fish-eye cameras face a strong distortion induced by the shape of the lens (see Figure~\ref{fig:undistortion}). Removing this distortion is a key preliminary step in many irradiance forecasting approaches such as cloud motion estimation~\cite{Modica2010, Quesada-Ruiz2014}. There exist two main approaches to tackle distortion.

First, one can make use of a transfer function to map distorted images into a more `readable' space given a model: a plane~\cite{Modica2010} or an icosahedral spherical polyhedron~\cite{Lee2019, Zhang2019} for instance. This is the most common strategy in traditional image-based solar forecasting. The alternative is to adapt the architecture of the model to a distorted input, which is an ongoing research field in Deep Learning with several publications in recent years~\cite{Eder2019, Cohen2019}.

\paragraph{Camera Callibration}
Finding the mapping between the distorted image and its planar projection can be achieved through a calibration of the camera. Assuming that the transfer function can be described by a Taylor series expansion, a proposed method consists of finding its coefficients through a four-step least-squares linear minimisation problem given the position of corner points of a chess board on a distorted image~\cite{Scaramuzza2006}. Alternatively, if the angular position of the Sun is known, its corresponding manually defined position in the image can be used to find a radial transfer function. This operation is therefore lens specific.

\paragraph{Azimuthal Equidistant Projection}
If such calibration data is not available, one can unwrap images by assuming the distortion induced by the fish-eye camera to be an azimuthal equidistant projection, which maintains angular distances~\cite{Bernecker2014}. The set of equations describing the transformation from polar ($\rho$, $\theta$) to spherical ($R$, $\phi$, $\psi$) coordinates are the following (Equations~\ref{equ:polar_spheric_coordinates_1} and~\ref{equ:polar_spheric_coordinates_2}):

\begin{equation}
  \cos \frac{\rho}{R} = \sin \phi_0 \: \sin \phi + \cos \phi_0 \: \cos \phi \: \cos (\psi - \psi_0)
  \label{equ:polar_spheric_coordinates_1}
\end{equation}

\begin{equation}
  \tan \theta = \frac{\cos \phi \: \sin(\psi - \psi_0)}{\cos \phi_0\:  \sin \phi - \sin \phi_0 \: \cos \phi \: \cos(\psi - \psi_0)}
  \label{equ:polar_spheric_coordinates_2}
\end{equation}

$\phi_0$ and $\psi_0$ define the reference axis, i.e. the normal to the ground in our application as the camera is facing upwards. $\psi_0$ is arbitrary so can thus be set to $0$, which simplifies equations~\ref{equ:polar_spheric_coordinates_1} and~\ref{equ:polar_spheric_coordinates_2} to Equations~\ref{equ:polar_spheric_coordinates_3}:

\begin{equation}
  \rho = R ( \frac{\pi}{2}-\phi ) \: \: \: \text{and} \: \: \: \theta = \psi
  \label{equ:polar_spheric_coordinates_3}
\end{equation}

Following this, the hemisphere is projected onto a plane by intersecting the different angular directions of the hemisphere with a horizontal square parallel to the ground. The resulting transformation is depicted in Figure~\ref{fig:undistortion}.

\newpage

\section*{Appendix B: Sun Trajectory}
\label{sun_trajectory}

\begin{figure}[H]
\centering    
\includegraphics[width=0.85\textwidth]{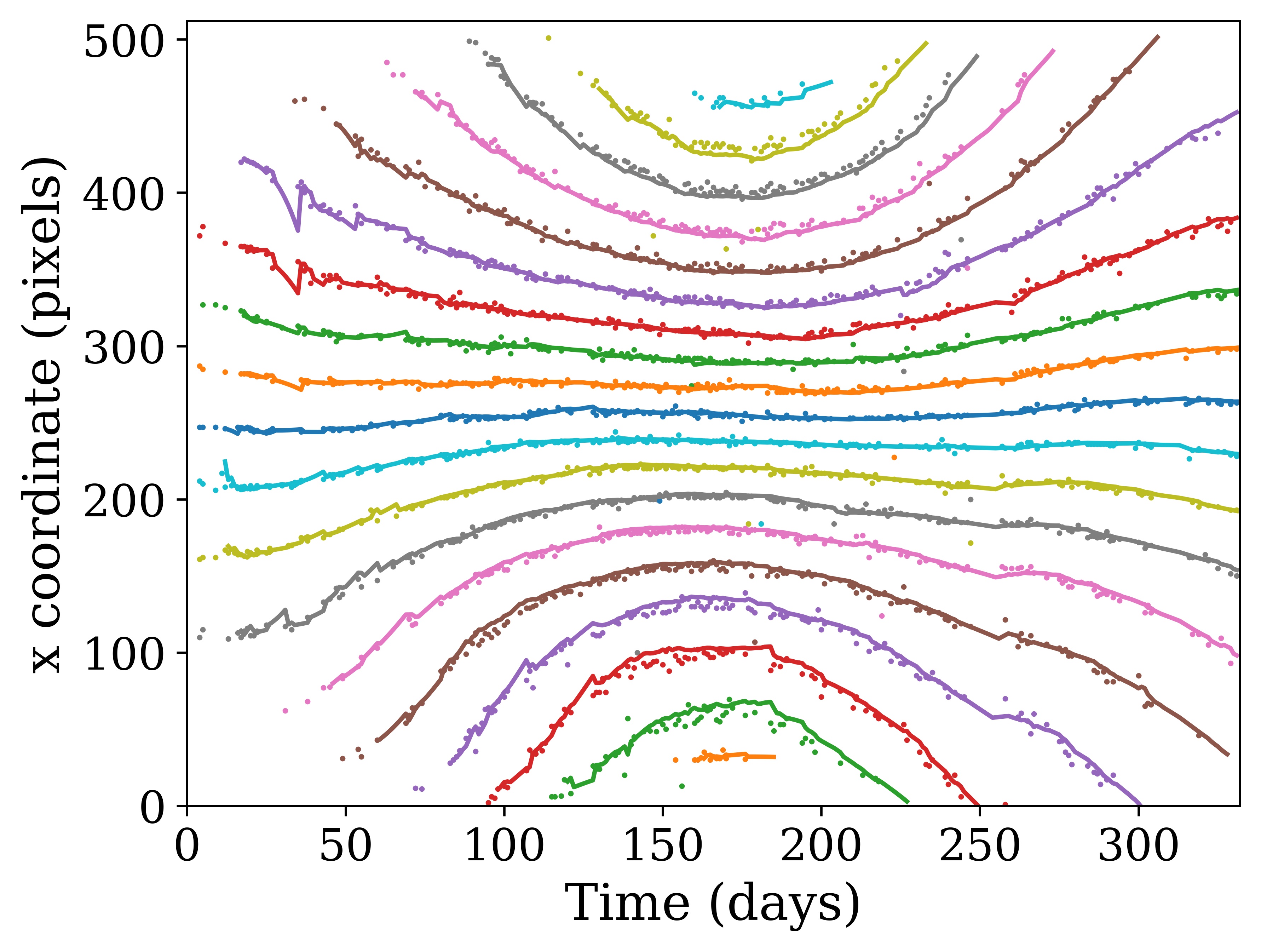}
\vspace{1\baselineskip}
\includegraphics[width=0.85\textwidth]{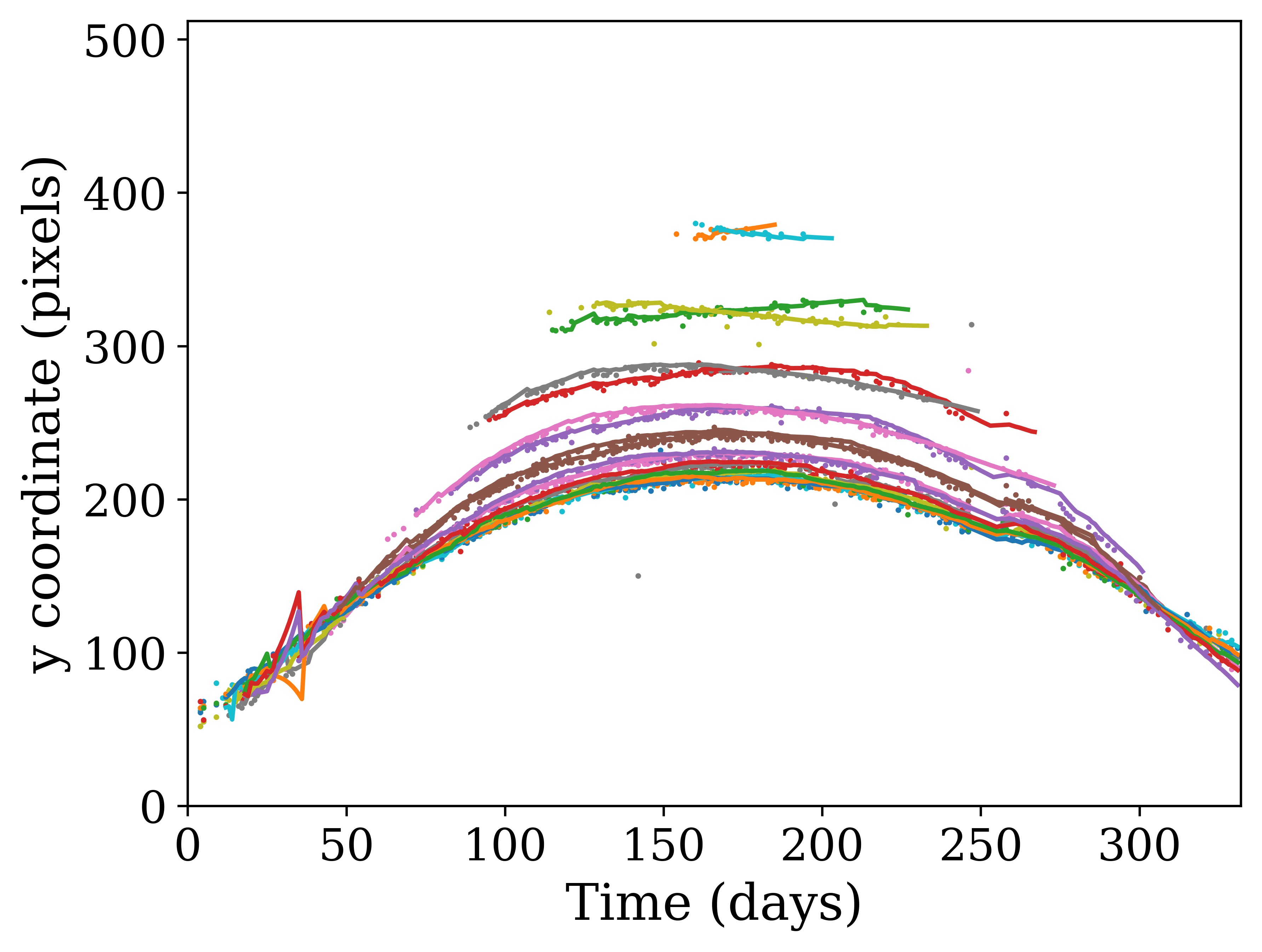}
\caption{\textit{x} and \textit{y} coordinates of the Sun for different minutes of the day along the year (points: position of the Sun when it is visible (before outlier removal), solid line: position estimates from previous days). The 19 curves of the Figure correspond to 19 minutes sampled every 25 min from 3:45 to 20:20}
\label{fig:sun_position_y_by_minute_19min}
\end{figure}

Figure~\ref{fig:sun_position_y_by_minute_19min} shows the trajectory of the Sun described by its coordinates \textit{x} and \textit{y} over a year. For every minute of the day, day-by-day coordinates (solid line) are interpolated from previous observations of the visible Sun represented by individual points in the figures. The largest errors occurring at the beginning of the year are due to the limited number of observations available to fit the model.

\section*{Appendix C: Example of Sun position predictions}
\label{examples}

\begin{figure}[ht]
  \centering
  
  \begin{minipage}[b]{0.37\textwidth}
    \includegraphics[width=\textwidth]{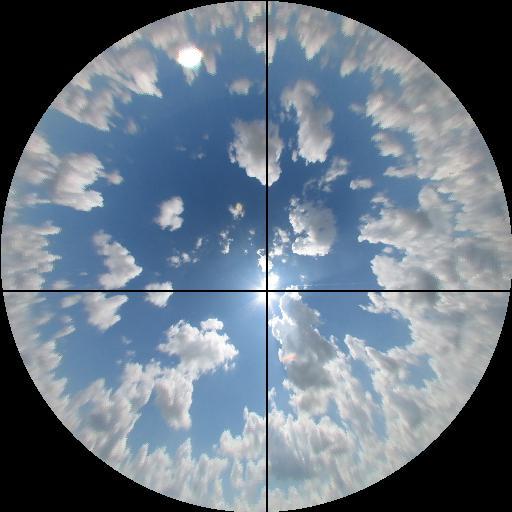}
    \label{fig:img_sh_tar_252_2018_7_3_12_32_visible}
  \end{minipage} 
  \quad \quad \quad \quad
  \begin{minipage}[b]{0.37\textwidth}
    \includegraphics[width=\textwidth]{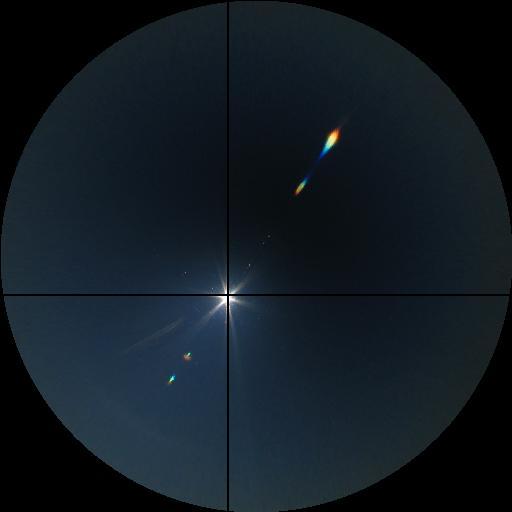}
    \label{fig:img_lg_tar_252_2018_8_2_10_46_pred}
  \end{minipage}
  
    \begin{minipage}[b]{0.37\textwidth}
    \includegraphics[width=\textwidth]{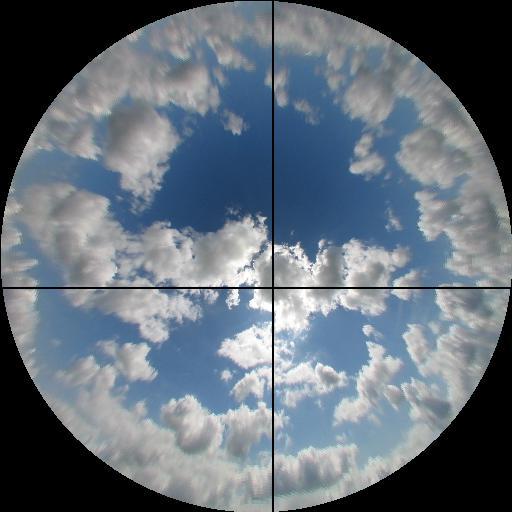}
    \label{fig:img_sh_tar_252_2018_6_2_12_52_pred}
  \end{minipage} 
  \quad \quad \quad \quad
  \begin{minipage}[b]{0.37\textwidth}
    \includegraphics[width=\textwidth]{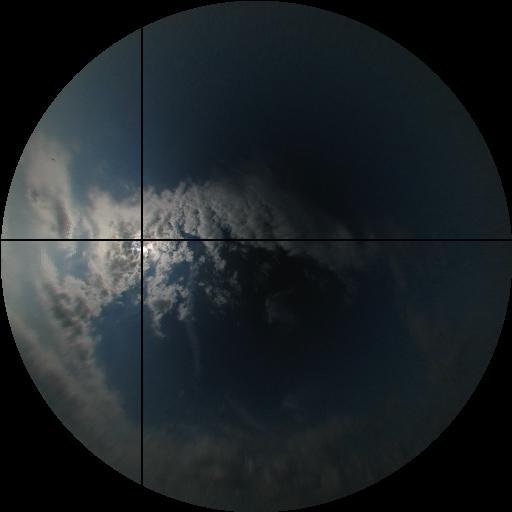}
    \label{fig:img_lg_tar_252_2018_6_3_6_58_pred}
  \end{minipage}
  
  \begin{minipage}[b]{0.37\textwidth}
    \includegraphics[width=\textwidth]{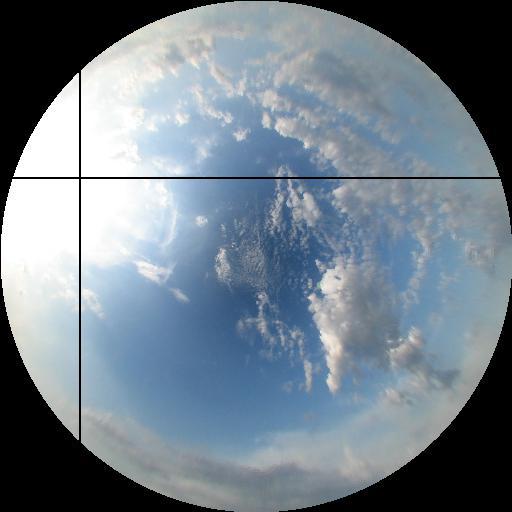}
    \label{fig:img_sh_tar_252_2018_7_2_5_28_pred}
  \end{minipage} 
  \quad \quad \quad \quad
  \begin{minipage}[b]{0.37\textwidth}
    \includegraphics[width=\textwidth]{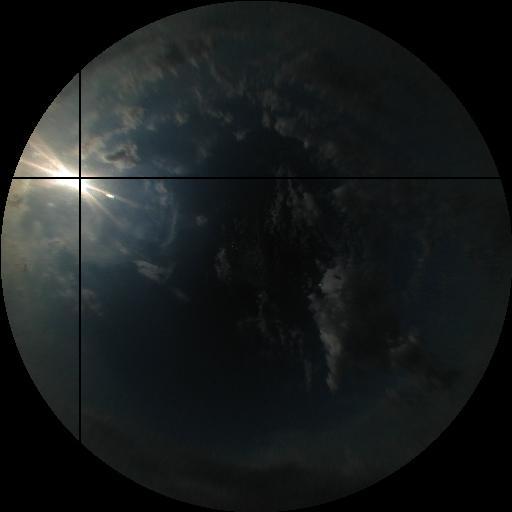}
    \label{fig:img_lg_tar_252_2018_7_2_5_28_pred}
  \end{minipage}
  \vspace{-0.5\baselineskip}
 \caption{Position of the Sun predicted by the Sun tracking algorithm from sky images (long exposures on the left and short exposures on the right). Top row: estimated position of the Sun when it is visible. The presence of flare in the right-hand image does not affect the prediction. Middle row: estimated position of the Sun when it is hidden or partially hidden by a cloud. Trajectory consistency provides reliable estimates of the Sun position. Bottom row: images of the sky taken simultaneously with different exposure times. In low Sun conditions the saturated area in the long exposure image makes the Sun localisation more challenging.}
 \label{fig:exmaples_3}
\end{figure}

Figure~\ref{fig:exmaples_3} presents qualitative results of the Sun tracking algorithm in different weather conditions and positions of the Sun on its daily east-west motion path. The origin shows the position of the Sun as predicted by the model.

\end{document}